\documentclass[11pt,a4paper]{article}
\usepackage[hyperref]{acl2017}
\usepackage{times}
\usepackage{latexsym}

\usepackage{url}
\usepackage{graphicx}
\usepackage{color}
\usepackage{amsmath}
\usepackage{bm}
\usepackage{booktabs}
\usepackage{multirow}
\usepackage{bm}
\usepackage{dblfloatfix}    
\usepackage{capt-of}
\usepackage{amssymb}
\usepackage{mathabx}
\usepackage [autostyle, english = american]{csquotes}
\usepackage{tabularx}
\usepackage{subcaption}
\usepackage{verbatim}

\usepackage{xcolor}
\usepackage{tikz}
\usepackage{pgfplots, pgfplotstable}
\usetikzlibrary{patterns}

\mathchardef\mhyphen="2D

\aclfinalcopy 
\def\aclpaperid{3249} 

\DeclareMathOperator*{\argmax}{arg\,max}
\DeclareMathOperator*{\topk}{topK}

\newcommand{\gap}{\vspace{2mm}}
\newcommand\given[1][]{\:#1\vert\:}
\newcommand{\code}[1]{\texttt{#1}}

\definecolor{gr}{HTML}{A6CEE3}
\definecolor{blu}{HTML}{1C9099}

\title{Data Augmentation for Low-Resource Neural Machine Translation}

\author{Marzieh Fadaee \qquad Arianna Bisazza \qquad Christof Monz\\
Informatics Institute, University of Amsterdam\\
Science Park 904, 1098 XH Amsterdam, The Netherlands\\
  {\tt \{m.fadaee,a.bisazza,c.monz\}@uva.nl} \\
 }

\date{}

\begin{document}
\maketitle
\begin{abstract}
The quality of a Neural Machine Translation system depends substantially on the availability of sizable parallel corpora.
For low-resource language pairs this is not the case, resulting in poor translation quality. 
Inspired by work in computer vision, we propose a novel data augmentation approach that targets low-frequency words by generating new sentence pairs containing rare words in new, synthetically created contexts.
Experimental results on simulated low-resource settings show that our method improves translation quality by up to 2.9 BLEU points over the baseline and up to 3.2 BLEU over back-translation.
\end{abstract}

\section{Introduction}

\begin{figure}[ht]
\centering
\includegraphics[width=\linewidth]{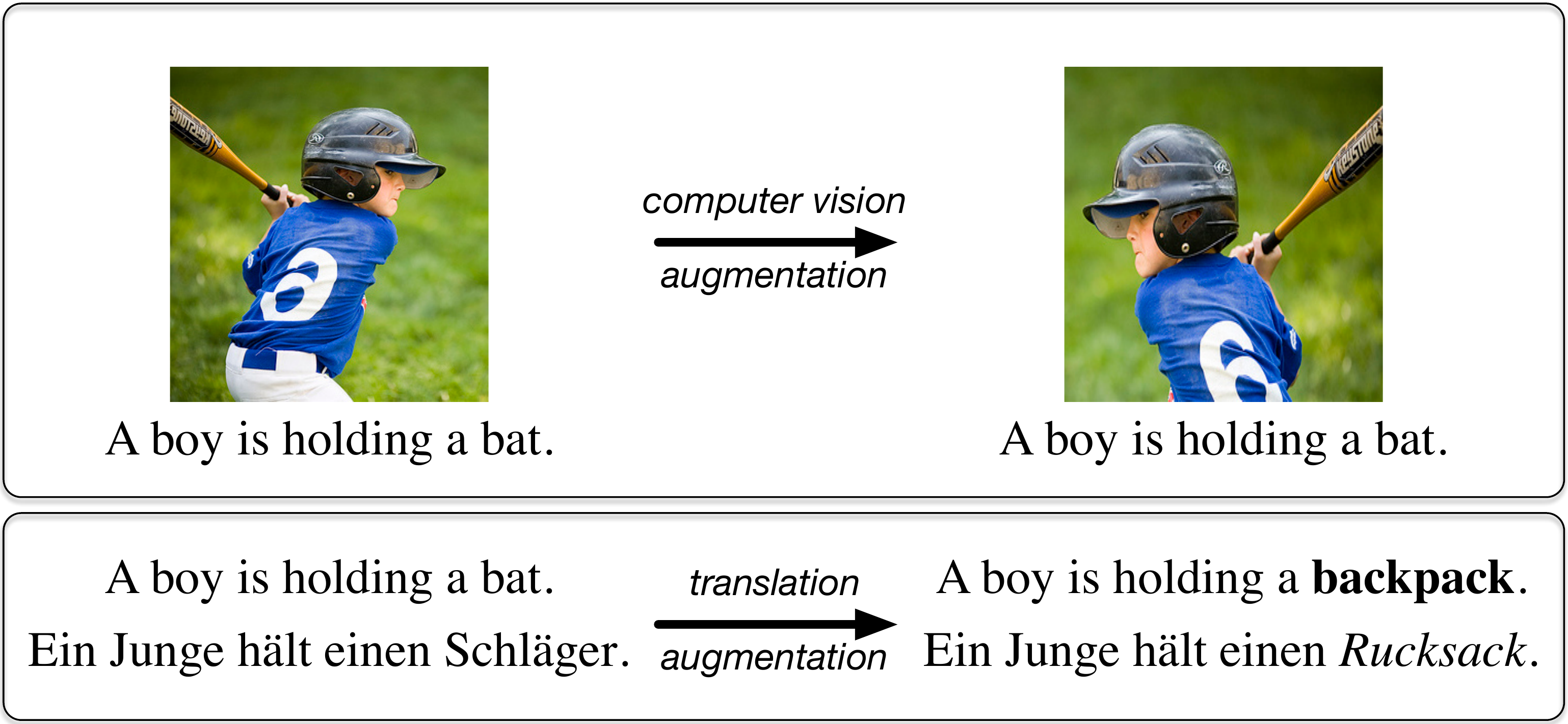}
\caption{Top: flip and crop, two label-preserving data augmentation techniques in computer vision.
Bottom: Altering one sentence in a parallel corpus requires changing its translation.}
\label{augfig}
\end{figure}

In computer vision, data augmentation techniques are widely used to increase robustness and improve learning of objects with a limited number of training examples. In image processing the training data is augmented  
by, for instance, horizontally flipping, random cropping, tilting, and altering the RGB channels of the original images \cite{NIPS2012_4824,DBLP:conf/bmvc/ChatfieldSVZ14}. Since the content of the new image is still the same, the label of the original image is preserved (see top of Figure~\ref{augfig}).
While data augmentation has become a standard technique to train deep networks for image processing, it is not a common practice in training networks for NLP tasks such as Machine Translation.

Neural Machine Translation (NMT) \cite{DBLP:journals/corr/BahdanauCB14,sutskever2014sequence,cho2014properties} 
is a sequence-to-sequence architecture 
where an encoder builds up a representation of the source sentence and a decoder, using the previous LSTM hidden states and an attention mechanism, generates the target translation. 

To train a model with reliable parameter estimations, these networks require numerous instances of sentence translation pairs with words occurring in diverse contexts, which is typically not available in low-resource language pairs.
As a result NMT falls short of reaching state-of-the-art performances for these language pairs \cite{zoph-EtAl:2016:EMNLP2016}.
The solution is to either manually annotate more data or perform unsupervised data augmentation. 
Since manual annotation of data is time-consuming, data augmentation for low-resource language pairs is a more viable approach.
Recently~\newcite{sennrich-haddow-birch:2016:P16-11} proposed a method to back-translate sentences from monolingual data and augment the bitext with the resulting pseudo parallel corpora. 

In this paper, we propose a simple yet effective approach, translation data augmentation (TDA), that augments the training data by altering existing sentences in the parallel corpus, similar in spirit to the data augmentation approaches in computer vision (see Figure~\ref{augfig}).
In order for the augmentation process in this scenario to be label-preserving, any change to a sentence in one language must preserve the meaning of the sentence, requiring sentential paraphrasing systems
which are not available for many language pairs. Instead, we propose a weaker notion of label preservation that allows to alter both source and target sentences at the same time as long as they remain translations of each other. 

While our approach allows us to augment data in numerous ways, we focus on augmenting instances involving low-frequency words, because the parameter estimation of rare words is challenging, and further exacerbated in a low-resource setting.
We simulate a low-resource setting as done in the literature \cite{Marton:2009:ISM:1699510.1699560,duong-EtAl:2015:EMNLP} and obtain substantial improvements for translating English$\rightarrow$German and German$\rightarrow$English.

\section{Translation Data Augmentation}

Given a source and target sentence pair (S,T), we want to alter it in a way that preserves the semantic equivalence between S and T while diversifying as much as possible the training examples.
A number of ways to do this can be envisaged, as for example paraphrasing (parts of) S or T.
Paraphrasing, however, is by itself a difficult task and is not guaranteed to bring useful new information into the training data.
We choose instead to focus on a subset of the vocabulary that we know to be poorly modeled by our baseline NMT system, namely words that occur rarely in the parallel corpus.
Thus, the goal of our data augmentation technique is to provide novel contexts for rare words.
To achieve this we search for contexts where a common word can be replaced by a rare word and consequently replace its corresponding word in the other language by that rare word's translation:
\vspace{1mm}

\begin{center}\small
\begin{tabular}{l l}
original pair & augmented pair \\
$S: s_1, ... , s_i, ..., s_n$ & $S': s_1, ... , s_i', ..., s_n$ \\
$T: t_1, ... , t_j, ..., t_m$ & $T': t_1, ... , t_j', ..., t_m$ \\ 
\end{tabular}
\end{center}

\vspace{1mm}

\noindent%
where $t_j$ is a translation of $s_i$ and word-aligned to $s_i$. 
Plausible substitutions are those that result in a fluent and grammatical sentence but do not necessarily maintain its semantic content.
As an example, the rare word \textit{motorbike} can be substituted in different contexts:

\smallskip

\begin{center}\small
\begin{tabular}{l | l}
Sentence [original / \textbf{substituted}] &  \begin{tabular}[x]{@{}c@{}}Plausible\end{tabular}  \\
\hline
	My sister drives a  [car / \textbf{motorbike}]  & yes \\
	My uncle sold his [house / \textbf{motorbike}] & yes \\
	Alice waters the [plant /  \textbf{motorbike}]  & no (semantics) \\
	John bought two [shirts / \textbf{motorbike}] & no (syntax) \\
\end{tabular}
\end{center}

\smallskip

\noindent%
Implausible substitutions need to be ruled out during data augmentation.
To this end, rather than relying on linguistic resources  
which are not available for many languages, 
we rely on LSTM language models (LM) \cite{Hochreiter1997,jozefowicz:2015} trained on large amounts of monolingual data in both forward and backward directions.

Our data augmentation method involves the following steps:
\paragraph{Targeted words selection:}
Following common practice, our NMT system limits its vocabulary $V$ to the $v$ most common words observed in the training corpus.
We select the words in $V$ that have fewer than $R$ occurrences and use this as our targeted rare word list $V_R$.
\paragraph{Rare word substitution:}
If the LM suggests a rare substitution in a particular context, we replace that word and add the new sentence to the training data.
Formally, given a sentence pair $(S,T)$ and a position $i$ in $S$ we compute the probability distribution over $V$ by the forward and backward LMs and select rare word substitutions $\mathcal{C}$ as follows:
\begin{align*}
\overrightarrow{\mathcal{C}} & = \{ s'_i \in V_{R} : \topk  P_{\textrm{\textit{ForwardLM}}_S}(s'_i \given s_{1}^{i-1}) \} \\
\overleftarrow{\mathcal{C}} & = \{ s'_i \in V_{R} : \topk P_{\textrm{\textit{BackwardLM}}_S}(s'_i \given s_{n}^{i+1}) \} \\
\mathcal{C} \ & = \{s'_i \given s'_i \in \overrightarrow{\mathcal{C}} \land s'_i \in \overleftarrow{\mathcal{C}} \} 
\end{align*}
where $\topk$ returns the $K$ words with highest conditional probability according to the context. 
The selected substitutions $s'_i$, are used to replace the original word and generate a new sentence.

\paragraph{Translation selection:}
Using automatic word alignments\footnote{We use fast-align \cite{dyer-chahuneau-smith:2013:NAACL-HLT} to extract word alignments and a bilingual lexicon with lexical translation probabilities from the low-resource bitext.} trained over the bitext, we replace the translation of word $s_i$ in $T$ by the translation of its substitution $s_i'$. 
Following a common practice in statistical MT, the optimal translation $t'_j$ is chosen by multiplying direct and inverse lexical translation probabilities with the LM probability of the translation in context: 
\begin{equation*}
t'_j =  \mathop{\argmax}_{t \in \textit{trans}(s_i')} P(s'_i \given t) P(t \given s'_i) P_{\textrm{\textit{LM}}_T}(t \given t_{1}^{j-1})
\end{equation*}
If no translation candidate is found because the word is unaligned or because
the LM probability is less than a certain threshold, the augmented sentence is discarded.
This reduces the risk of generating sentence pairs that are semantically or syntactically incorrect.
\paragraph{Sampling:}
We loop over the original parallel corpus multiple times, sampling substitution positions, $i$, in each sentence and making sure that each rare word gets augmented at most $N$ times so that a large number of rare words can be affected.
We stop when no new sentences are generated in one pass of the training data.

\gap
Table~\ref{examples} provides some examples resulting from our augmentation procedure. While using a large LM to substitute words with rare words mostly results in grammatical sentences, this does not mean that the meaning of the original sentence is preserved. Note that meaning preservation is not an objective of our approach.  

Two translation data augmentation (TDA) setups are considered: only one word per sentence can be replaced (TDA$_{r=1}$),  or multiple words per sentence can be replaced, with the condition that any two replaced words are at least five positions apart (TDA$_{r\ge1}$).
The latter incurs a higher risk of introducing noisy sentences but has the potential to positively affect more rare words within the same amount of augmented data. 
We evaluate both setups in the following section.

\begin{table}[htb!]
\centering
\small
\begin{tabularx}{\linewidth}{@{\ }l @{:\ \ \ }X@{\ }}
 \toprule
 En & I had been told that you would [not / \textbf{voluntarily}] be speaking today. \\
De & mir wurde signalisiert, sie w{\"u}rden heute [nicht / \textit{freiwillig}] sprechen.\\
\midrule
 En & the present situation is [indefensible / \textbf{confusing}] and completely unacceptable to the commission.\\
 De & die situation sei [unhaltbar / \textit{verwirrend}] und f{\"u}r die kommission g{\"a}nzlich unannehmbar.\\
\midrule
 En & ... agree wholeheartedly with the institution of an ad hoc delegation of parliament on the turkish [prison / \textbf{missile}] system.\\
 De &  ... ad-hoc delegation des parlaments f{\"u}r das regime in den t{\"u}rkischen [gef{\"a}ngnissen / \textit{flugwaffen}] voll und ganz zustimmen.\\
\bottomrule
\end{tabularx}
\caption{Examples of augmented data with highlighted [original / \textbf{substituted}] and [original / \textit{translated}] words.
}
\label{examples}
\end{table}

\newcommand{\sigspace}{\rule{3ex}{0pt}}
\newcommand{\sigspacehalf}{\rule{1.6ex}{0pt}}
\begin{table*}[ht!]
\centering
\small%
\tabcolsep=2.5pt%
\begin{tabular}{@{\ }l|r|rlrlrl|rlrlrl@{}}
\multicolumn{2}{c}{}    & \multicolumn{6}{c}{De-En} & \multicolumn{6}{c}{En-De}  \\\hline
Model   & Data &  \multicolumn{2}{c}{test2014} &  \multicolumn{2}{c}{test2015} &  \multicolumn{2}{c|}{test2016} &  \multicolumn{2}{c}{test2014} &  \multicolumn{2}{c}{test2015} &  \multicolumn{2}{c}{test2016}\\ \hline     Full data (ceiling)  & 3.9M &  21.1 &  & 22.0 & & 26.9 & & 17.0 & & 18.5 & & 21.7 & \\\hline
Baseline  & 371K & 10.6 & & 11.3 & & 13.1 & & 8.2 & & 9.2 & & 11.0 & \\
Back-translation$_{1:1}$  & 731K & 11.4 & (+0.8)$^{\blacktriangleup}$\sigspacehalf & 12.2 & (+0.9)$^{\blacktriangleup}$\sigspacehalf & 14.6 & (+1.5)$^{\blacktriangleup}$\sigspacehalf & 9.0 & (+0.8)$^{\blacktriangleup}$\sigspacehalf & 10.4 & (+1.2)$^{\blacktriangleup}$\sigspacehalf & 12.0 & (+1.0)$^{\blacktriangleup}\sigspacehalf$ \\
Back-translation$_{3:1}$  & 1.5M & 11.2 & (+0.6)\sigspace & 11.2 & (--0.1)\sigspace & 13.3 & (+0.2)\sigspace& 7.8 & (--0.4)\sigspace & 9.4 & (+0.2)\sigspace & 10.7 & (--0.3)\sigspace \\\hline
TDA$_{r=1}$ & 4.5M  &  11.9 & (+1.3)$^{\blacktriangleup,\mhyphen}$ & 13.4 & (+2.1)$^{\blacktriangleup,\blacktriangleup}$ & 15.2 & (+2.1)$^{\blacktriangleup,\blacktriangleup}$  & 10.4 & (+2.2)$^{\blacktriangleup,\blacktriangleup}$  & 11.2 & (+2.0)$^{\blacktriangleup,\blacktriangleup}$ & 13.5 & (+2.5)$^{\blacktriangleup,\blacktriangleup}$  \\
TDA$_{r\ge 1}$ &  6M    &  \textbf{12.6}  & (+2.0)$^{\blacktriangleup,\blacktriangleup}$ & \textbf{13.7} & (+2.4)$^{\blacktriangleup,\blacktriangleup}$ & \textbf{15.4} & (+2.3)$^{\blacktriangleup,\blacktriangleup}$  & \textbf{10.7} & (+2.5)$^{\blacktriangleup,\blacktriangleup}$  & \textbf{11.5} & (+2.3)$^{\blacktriangleup,\blacktriangleup}$ & \textbf{13.9} & (+2.9)$^{\blacktriangleup,\blacktriangleup}$ \\
Oversampling & 6M & 11.9 & (+1.3)$^{\blacktriangleup,\mhyphen}$ & 12.9 & (+1.6)$^{\blacktriangleup,\smalltriangleup}$ & 15.0 & (+1.9)$^{\blacktriangleup,\mhyphen}$ & 9.7 & (+1.5)$^{\blacktriangleup,\smalltriangleup}$ & 10.7 & (+1.5)$^{\blacktriangleup,\mhyphen}$ & 12.6 & (+1.6)$^{\blacktriangleup,\mhyphen}$ \\\hline
\end{tabular}
\caption{\label{bleuTB}Translation performance (BLEU) on German-English and English-German WMT test sets (newstest2014, 2015, and 2016) in a simulated low-resource setting. Back-translation refers to the work of~\newcite{sennrich-haddow-birch:2016:P16-11}. 
Statistically significant improvements are marked $^\blacktriangleup$ at the $p < .01$ and $^\smalltriangleup$ at the $p < .05$ level, with the first superscript referring to baseline and the second to back-translation$_{1:1}$.}
\end{table*}

\section{Evaluation}

In this section we evaluate the utility of our approach in a simulated low-resource NMT scenario.

\subsection{Data and experimental setup}

To simulate a low-resource setting we randomly sample 10\% of the English$\leftrightarrow$German WMT15 training data and report results on newstest 2014, 2015, and 2016 \cite{bojar-EtAl:2016:WMT1}. 
For reference we also provide the result of our baseline system on the full data.

As NMT system we use a 4-layer attention-based encoder-decoder model as described in \cite{luong:2015:EMNLP} trained with hidden dimension 1000, batch size 80 for 20 epochs.
In all experiments the NMT vocabulary is limited to the most common 30K words in both languages.
Note that data augmentation does not introduce new words to the vocabulary.
In all experiments we preprocess source and target language data with Byte-pair encoding (BPE) \cite{sennrich-haddow-birch:2016:P16-12} using 30K merge operations. 
In the augmentation experiments BPE is performed after data augmentation. 

For the LMs needed for data augmentation, we train 2-layer LSTM networks in forward and backward directions on the monolingual data provided for the same task 
(3.5B and 0.9B tokens in English and German respectively) 
with embedding size 64 and hidden size 128.
We set the rare word threshold $R$ to 100, top $K$ words to 1000 and maximum number $N$ of augmentations per rare word to 500.
In all experiments we use the English LM for the rare word substitutions, and the German LM to choose the optimal word translation in context.
Since our approach is not label preserving we only perform augmentation during training and do not alter source sentences during testing.

We also compare our approach to~\newcite{sennrich-haddow-birch:2016:P16-11} by back-translating monolingual data and adding it to the parallel training data.
Specifically, we back-translate sentences from the target side of WMT'15 that are not included in our low-resource baseline with two settings: keeping a one-to-one ratio of back-translated  
versus original data (\mbox{$1:1$}) following the authors' suggestion, or using three times more back-translated data (\mbox{$3:1$}).

We measure translation quality by single-reference case-insensitive BLEU \cite{Papineni2001} computed with the \code{multi-bleu.perl} script from Moses.

\subsection{Results}

All translation results are displayed in Table~\ref{bleuTB}.
As expected, the low-resource baseline performs much worse than the full data system,
re-iterating the importance of sizable training data for NMT. 
Next we observe that both back-translation and our proposed TDA method significantly improve translation quality. However TDA obtains the best results overall and significantly outperforms back-translation in all test sets. 
This is an important finding considering that our method involves only minor modifications to the original training sentences and does not involve any costly translation process. 
Improvements are consistent across both translation directions,  
regardless of whether rare word substitutions are first applied to the source or to the target side.

We also observe that altering multiple words in a sentence performs slightly better than altering only one.
This indicates that addressing more rare words is preferable even though the augmented sentences are likely to be noisier.  

\begin{table*}[htb!]
\centering
\small
\begin{tabularx}{\linewidth}{@{\ }l @{\ \ \ }X@{\ }}
 \toprule
Source & der tunnel hat einen querschnitt von 1,20 meter h{\"o}he und 90 zentimeter breite .\\ 
Baseline translation & the wine consists of about 1,20 m and 90 of the canal . \\
TDA\textsubscript{$_{r\ge 1}$} translation & the tunnel has a UNK measuring meters 1.20 metres high and 90 \textbf{centimetres} wide . \\
Reference  & the tunnel has a cross - section measuring 1.20 metres high and 90 centimetres across . \\
\midrule
 Examples of & $\bullet$ the average speed of cars and buses is therefore around 20 [kilometres / \textbf{centimetres}]  per hour .\\ 
augmented data &  $\bullet$ grab crane in special terminals for handling capacities of up to 1,800 [tonnes / \textbf{centimetres}] per hour .\\ 
for the word & $\bullet$ all suites and rooms are very spacious and measure between 50 and 70 [m / \textbf{centimetres}] \\ 
 \textit{centimetres} & $\bullet$ all we have to do is lower the speed limit everywhere to five [kilometers / \textbf{centimetres}]  per hour .\\ 
\bottomrule
\end{tabularx}
\caption{An example from newstest2014 illustrating the effect of augmenting rare words on generation during test time. The translation of the baseline does not include the rare word \textit{centimetres}, however, the translation of our TDA model generates the rare word and produces a more fluent sentence. Instances of the augmentation of the word \textit{centimetres} in training data are also provided.}
\label{transex}
\end{table*}

To verify that the gains are actually due to the rare word substitutions and not just to the repetition of part of the training data, we perform a final experiment where each sentence pair selected for augmentation is added to the training data \textit{unchanged} (Oversampling in Table~\ref{bleuTB}). 
Surprisingly, we find that this simple form of sampled data replication outperforms both baseline and back-translation systems,\footnote{Note that this effect cannot be achieved by simply continuing the baseline training for up to 50 epochs.} while TDA\textsubscript{$_{r\ge 1}$} remains the best performing system overall.

We also observe that the system trained on augmented data tends to generate longer translations. 
Averaging on all test sets, the length of translations generated by the baseline is 0.88 of the average reference length, while for TDA\textsubscript{$_{r=1}$} and TDA\textsubscript{$_{r\ge 1}$} it is 0.95 and 0.94, respectively.
We attribute this effect to the ability of the TDA-trained system to generate translations for rare words that were left untranslated by the baseline system.

\section{Analysis of the Results}

A desired effect of our method is to increase the number of correct rare words generated by the NMT system at test time. 

To examine the impact of augmenting the training data by creating contexts for rare words on the target side, Table~\ref{transex} provides an example for German$\rightarrow$English translation. We see that the baseline model is not able to generate the rare word \textit{centimetres} as a correct translation of the German word \textit{zentimeter }. However, this word is not rare in the training data of the TDA\textsubscript{$_{r\ge 1}$} model after augmentation and is generated during translation. Table~\ref{transex} also provides several instances of augmented training sentences targeting the word \textit{centimetres}. Note that even though some augmented sentences are nonsensical (e.g. \textit{the speed limit is five centimetres per hour}), the NMT system still benefits from the new context for the rare word and is able to generate it during testing.

\begin{figure}[ht]
\centering
\begin{tikzpicture}[shorten >=1pt, node distance=2.5cm,scale=0.8]
\pgfplotstableread{
X   Gp   Name        generated     rare         notrareanymoreGen   notrareanymoreNotGen 
8   baseline   14    604   3389  0   0
7   TDA test 2114   3389     2082  32   
5   baseline   15    480   2528    0    0
4   TDA   test    1522   2528    1422    100
2   baseline   16    627   3146   0    0
1   TDA   test  2072   3146    1998  70
}\datatable
\begin{axis}[
     width=8.5cm, height=6.5cm,
     xmin=0,
     xbar,
     xmax = 3500, 
     bar width=13pt,
     bar shift=0pt,
     ytick=data,
     reverse legend,
         xtick={1000, 2000, 3000},
     yticklabels from table={\datatable}{Gp},
         ytick style={draw=none},
     legend style={at={(0.42,-0.15)},anchor=north, draw=none, /tikz/every even column/.append style={column sep=5pt}}, 
     legend image code/.code={%
             \draw[#1] (0cm,-0.1cm) rectangle (0.3cm,0.1cm);
        },
     legend cell align=left
]

\addplot [fill=blu, fill opacity=1] table [y=X, x=rare] {\datatable};  \addlegendentry{{Words in $V_R \cap V_{ref}$ not generated during translation}} 
\addplot [fill=gr, fill opacity=1] table [y=X, x=generated] {\datatable}; \addlegendentry{{Words in $V_R \cap V_{ref}$ generated during translation }}
\end{axis}

\begin{axis}[
    width=8.5cm, height=6.5cm,
    xmin=0,
    xbar stacked,
    xmax = 3000,
    bar width=13pt,
    ytick=data,
    axis y line*=right,
    y tick label style={rotate=90} ,
     yticklabels from table={\datatable}{Name},
    xticklabels=\empty,
    xtick=\empty,
    ytick style={draw=none},
     legend style={at={(0.35,-0.345)},anchor=north, draw=none, /tikz/every even column/.append style={column sep=5pt}}, 
     legend image code/.code={%
             \draw [pattern=north east lines](0cm,-0.1cm) rectangle (0.3cm,0.1cm);
        },
     legend cell align=left,
]
\addplot [fill=red, fill opacity=0,forget plot] table [y=X, x=notrareanymoreNotGen] {\datatable};
\addplot [pattern=north east lines] table [y=X, x=notrareanymoreGen] {\datatable}; \addlegendentry{Words in $V_R \cap V_{ref}$ affected by augmentation}

\end{axis}
\end{tikzpicture}
\caption{Effect of TDA on the number of unique rare words generated during De$\rightarrow$En translation. $V_R$ is the set of rare words targeted by TDA\textsubscript{$_{r\ge 1}$} and $V_{ref}$ the reference translation vocabulary.} 
\label{rarewordfreqs}
\end{figure}
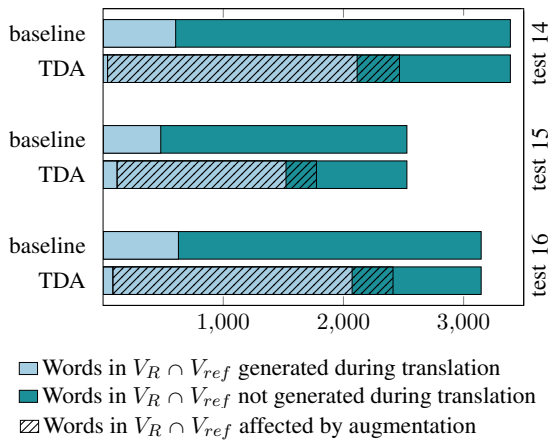

Figure~\ref{rarewordfreqs} demonstrates that this is indeed the case for many words: the number of rare words occurring in the reference translation ($V_R \cap V_{ref}$) is three times larger in the TDA system output than in the baseline output. 
One can also see that this increase is a direct effect of TDA
as most of the rare words are not `rare' anymore in the augmented data, i.e., they were augmented sufficiently many times to occur more than 100 times (see hatched pattern in Figure~\ref{rarewordfreqs}).
Note that during the experiments we did not use any information from the evaluation sets.

To gauge the impact of augmenting the contexts for rare words on the source side, we examine normalized attention scores of these words before and after augmentation. When translating English$\rightarrow$German 
with our TDA model, the attention scores for rare words on the source side are on average 8.8\% higher than when translating with the baseline model. This suggests that having more accurate representations of rare words increases the model's confidence to attend to these words when encountered during test time.

\begin{table}[htb!]
\centering
\small
\begin{tabularx}{\linewidth}{@{\ }l @{:\ \ \ }X@{\ }}
 \toprule
En & registered users will receive the UNK newsletter free [of / \textbf{yearly}] charge.\\
De & registrierte user erhalten zudem regelm{\"a}{\ss}ig [den / \textit{j{\"a}hrlich}] markenticker newsletter.\\
\midrule
 En & the personal contact is [essential / \textbf{entrusted}] to us\\
 De &  pers{\"o}nliche kontakt ist uns sehr [wichtig / \textit{betraut}] \\
\bottomrule
\end{tabularx}
\caption{Examples of incorrectly augmented data with highlighted [original / \textbf{substituted}] and [original / \textit{translated}] words.
}
\label{examplesinc}
\end{table}

Finally Table~\ref{examplesinc} provides examples of cases where augmentation results in incorrect sentences. 
In the first example, the sentence is ungrammatical after substitution (\textit{of / yearly}), which can be the result of choosing substitutions with low probabilities from the English LM $\topk$ suggestions. 

Errors can also occur during translation selection, as in the second example where \textit{betraut} is an acceptable translation of \textit{entrusted} but would require a rephrasing of the German sentence to be grammatically correct.
Problems of this kind can be attributed to the German LM, but also to the lack of a more suitable translation in the lexicon extracted from the bitext.
Interestingly, this noise seems to affect NMT only to a limited extent.

\section{Conclusion}

We have proposed a simple but effective approach to augment the training data of Neural Machine Translation for low-resource language pairs. 
By leveraging language models trained on large amounts of monolingual data,
we generate new sentence pairs containing rare words in new, synthetically created contexts. 
We show that this approach leads to generating more rare words during translation and, consequently, to higher translation quality. 
In particular we report substantial improvements in simulated low-resource English$\rightarrow$German and German$\rightarrow$English settings, outperforming another recently proposed data augmentation technique.

\section*{Acknowledgments}
This research was funded in part by the Netherlands Organization for Scientific Research (NWO) under project numbers 639.022.213 and 639.021.646, and a Google Faculty Research Award. We also thank NVIDIA for their hardware support, Ke Tran for providing the neural machine translation baseline system, and the anonymous reviewers for their helpful comments. 

\bibliography{acl2017}

\begin{thebibliography}{}
\expandafter\ifx\csname natexlab\endcsname\relax\def\natexlab#1{#1}\fi

\bibitem[{Bahdanau et~al.(2015)Bahdanau, Cho, and
  Bengio}]{DBLP:journals/corr/BahdanauCB14}
Dzmitry Bahdanau, Kyunghyun Cho, and Yoshua Bengio. 2015.
\newblock Neural machine translation by jointly learning to align and
  translate.
\newblock In {\em Proceedings of the International Conference on Learning
  Representations (ICLR)\/}.

\bibitem[{Bojar et~al.(2016)Bojar, Chatterjee, Federmann, Graham, Haddow, Huck,
  Jimeno~Yepes, Koehn, Logacheva, Monz, Negri, Neveol, Neves, Popel, Post,
  Rubino, Scarton, Specia, Turchi, Verspoor, and
  Zampieri}]{bojar-EtAl:2016:WMT1}
Ond\v{r}ej Bojar, Rajen Chatterjee, Christian Federmann, Yvette Graham, Barry
  Haddow, Matthias Huck, Antonio Jimeno~Yepes, Philipp Koehn, Varvara
  Logacheva, Christof Monz, Matteo Negri, Aurelie Neveol, Mariana Neves, Martin
  Popel, Matt Post, Raphael Rubino, Carolina Scarton, Lucia Specia, Marco
  Turchi, Karin Verspoor, and Marcos Zampieri. 2016.
\newblock \href{http://www.aclweb.org/anthology/W/W16/W16-2301}{Findings of the
  2016 conference on machine translation}.
\newblock In {\em Proceedings of the First Conference on Machine
  Translation\/}. Association for Computational Linguistics, Berlin, Germany,
  pages 131--198.
\newblock
  \href{http://www.aclweb.org/anthology/W/W16/W16-2301}{http://www.aclweb.org/anthology/W/W16/W16-2301}.

\bibitem[{Chatfield et~al.(2014)Chatfield, Simonyan, Vedaldi, and
  Zisserman}]{DBLP:conf/bmvc/ChatfieldSVZ14}
Ken Chatfield, Karen Simonyan, Andrea Vedaldi, and Andrew Zisserman. 2014.
\newblock \href{https://doi.org/http://dx.doi.org/10.5244/C.28.6}{Return of the
  devil in the details: Delving deep into convolutional nets}.
\newblock In {\em Proceedings of the British Machine Vision Conference\/}. BMVA
  Press.
\newblock
  \href{https://doi.org/http://dx.doi.org/10.5244/C.28.6}{https://doi.org/http://dx.doi.org/10.5244/C.28.6}.

\bibitem[{Cho et~al.(2014)Cho, {van Merrienboer}, Bahdanau, and
  Bengio}]{cho2014properties}
Kyunghyun Cho, B~{van Merrienboer}, Dzmitry Bahdanau, and Yoshua Bengio. 2014.
\newblock On the properties of neural machine translation: Encoder-decoder
  approaches.
\newblock In {\em Eighth Workshop on Syntax, Semantics and Structure in
  Statistical Translation (SSST-8), 2014\/}.

\bibitem[{Duong et~al.(2015)Duong, Cohn, Bird, and
  Cook}]{duong-EtAl:2015:EMNLP}
Long Duong, Trevor Cohn, Steven Bird, and Paul Cook. 2015.
\newblock \href{http://aclweb.org/anthology/D15-1040}{A neural network model
  for low-resource universal dependency parsing}.
\newblock In {\em Proceedings of the 2015 Conference on Empirical Methods in
  Natural Language Processing\/}. Association for Computational Linguistics,
  Lisbon, Portugal, pages 339--348.
\newblock
  \href{http://aclweb.org/anthology/D15-1040}{http://aclweb.org/anthology/D15-1040}.

\bibitem[{Dyer et~al.(2013)Dyer, Chahuneau, and
  Smith}]{dyer-chahuneau-smith:2013:NAACL-HLT}
Chris Dyer, Victor Chahuneau, and Noah~A. Smith. 2013.
\newblock \href{http://www.aclweb.org/anthology/N13-1073}{A simple, fast, and
  effective reparameterization of ibm model 2}.
\newblock In {\em Proceedings of the 2013 Conference of the North American
  Chapter of the Association for Computational Linguistics: Human Language
  Technologies\/}. Association for Computational Linguistics, Atlanta, Georgia,
  pages 644--648.
\newblock
  \href{http://www.aclweb.org/anthology/N13-1073}{http://www.aclweb.org/anthology/N13-1073}.

\bibitem[{Hochreiter and Schmidhuber(1997)}]{Hochreiter1997}
Sepp Hochreiter and J\"{u}rgen Schmidhuber. 1997.
\newblock Long short-term memory.
\newblock {\em Neural Computation\/} 9(8):1735--1780.

\bibitem[{Jozefowicz et~al.(2015)Jozefowicz, Zaremba, and
  Sutskever}]{jozefowicz:2015}
Rafal Jozefowicz, Wojciech Zaremba, and Ilya Sutskever. 2015.
\newblock \href{http://proceedings.mlr.press/v37/jozefowicz15.pdf}{An empirical
  exploration of recurrent network architectures}.
\newblock In Francis Bach and David Blei, editors, {\em Proceedings of the 32nd
  International Conference on Machine Learning\/}. PMLR, Lille, France,
  volume~37 of {\em Proceedings of Machine Learning Research\/}, pages
  2342--2350.
\newblock
  \href{http://proceedings.mlr.press/v37/jozefowicz15.pdf}{http://proceedings.mlr.press/v37/jozefowicz15.pdf}.

\bibitem[{Krizhevsky et~al.(2012)Krizhevsky, Sutskever, and
  Hinton}]{NIPS2012_4824}
Alex Krizhevsky, Ilya Sutskever, and Geoffrey~E Hinton. 2012.
\newblock
  \href{http://papers.nips.cc/paper/4824-imagenet-classification-with-deep-convolutional-neural-networks.pdf}{Imagenet
  classification with deep convolutional neural networks}.
\newblock In F.~Pereira, C.~J.~C. Burges, L.~Bottou, and K.~Q. Weinberger,
  editors, {\em Advances in Neural Information Processing Systems 25\/}, Curran
  Associates, Inc., pages 1097--1105.
\newblock
  \href{http://papers.nips.cc/paper/4824-imagenet-classification-with-deep-convolutional-neural-networks.pdf}{http://papers.nips.cc/paper/4824-imagenet-classification-with-deep-convolutional-neural-networks.pdf}.

\bibitem[{Luong et~al.(2015)Luong, Pham, and Manning}]{luong:2015:EMNLP}
Thang Luong, Hieu Pham, and Christopher~D. Manning. 2015.
\newblock \href{http://aclweb.org/anthology/D15-1166}{Effective approaches to
  attention-based neural machine translation}.
\newblock In {\em Proceedings of the 2015 Conference on Empirical Methods in
  Natural Language Processing\/}. Association for Computational Linguistics,
  Lisbon, Portugal, pages 1412--1421.
\newblock
  \href{http://aclweb.org/anthology/D15-1166}{http://aclweb.org/anthology/D15-1166}.

\bibitem[{Marton et~al.(2009)Marton, Callison-Burch, and
  Resnik}]{Marton:2009:ISM:1699510.1699560}
Yuval Marton, Chris Callison-Burch, and Philip Resnik. 2009.
\newblock \href{http://www.aclweb.org/anthology/D/D09/D09-1040}{Improved
  statistical machine translation using monolingually-derived paraphrases}.
\newblock In {\em Proceedings of the 2009 Conference on Empirical Methods in
  Natural Language Processing\/}. Association for Computational Linguistics,
  Singapore, pages 381--390.
\newblock
  \href{http://www.aclweb.org/anthology/D/D09/D09-1040}{http://www.aclweb.org/anthology/D/D09/D09-1040}.

\bibitem[{Papineni et~al.(2002)Papineni, Roukos, Ward, and Zhu}]{Papineni2001}
Kishore Papineni, Salim Roukos, Todd Ward, and Wei-Jing Zhu. 2002.
\newblock \href{https://doi.org/10.3115/1073083.1073135}{Bleu: a method for
  automatic evaluation of machine translation}.
\newblock In {\em Proceedings of 40th Annual Meeting of the Association for
  Computational Linguistics\/}. Association for Computational Linguistics,
  Philadelphia, Pennsylvania, USA, pages 311--318.
\newblock
  \href{https://doi.org/10.3115/1073083.1073135}{https://doi.org/10.3115/1073083.1073135}.

\bibitem[{Sennrich et~al.(2016{\natexlab{a}})Sennrich, Haddow, and
  Birch}]{sennrich-haddow-birch:2016:P16-11}
Rico Sennrich, Barry Haddow, and Alexandra Birch. 2016{\natexlab{a}}.
\newblock \href{http://www.aclweb.org/anthology/P16-1009}{Improving neural
  machine translation models with monolingual data}.
\newblock In {\em Proceedings of the 54th Annual Meeting of the Association for
  Computational Linguistics (Volume 1: Long Papers)\/}. Association for
  Computational Linguistics, Berlin, Germany, pages 86--96.
\newblock
  \href{http://www.aclweb.org/anthology/P16-1009}{http://www.aclweb.org/anthology/P16-1009}.

\bibitem[{Sennrich et~al.(2016{\natexlab{b}})Sennrich, Haddow, and
  Birch}]{sennrich-haddow-birch:2016:P16-12}
Rico Sennrich, Barry Haddow, and Alexandra Birch. 2016{\natexlab{b}}.
\newblock \href{http://www.aclweb.org/anthology/P16-1162}{Neural machine
  translation of rare words with subword units}.
\newblock In {\em Proceedings of the 54th Annual Meeting of the Association for
  Computational Linguistics (Volume 1: Long Papers)\/}. Association for
  Computational Linguistics, Berlin, Germany, pages 1715--1725.
\newblock
  \href{http://www.aclweb.org/anthology/P16-1162}{http://www.aclweb.org/anthology/P16-1162}.

\bibitem[{Sutskever et~al.(2014)Sutskever, Vinyals, and
  Le}]{sutskever2014sequence}
Ilya Sutskever, Oriol Vinyals, and Quoc~V Le. 2014.
\newblock
  \href{http://papers.nips.cc/paper/5346-sequence-to-sequence-learning-with-neural-networks.pdf}{Sequence
  to sequence learning with neural networks}.
\newblock In Z.~Ghahramani, M.~Welling, C.~Cortes, N.~D. Lawrence, and K.~Q.
  Weinberger, editors, {\em Advances in Neural Information Processing Systems
  27\/}, Curran Associates, Inc., pages 3104--3112.
\newblock
  \href{http://papers.nips.cc/paper/5346-sequence-to-sequence-learning-with-neural-networks.pdf}{http://papers.nips.cc/paper/5346-sequence-to-sequence-learning-with-neural-networks.pdf}.

\bibitem[{Zoph et~al.(2016)Zoph, Yuret, May, and
  Knight}]{zoph-EtAl:2016:EMNLP2016}
Barret Zoph, Deniz Yuret, Jonathan May, and Kevin Knight. 2016.
\newblock \href{https://aclweb.org/anthology/D16-1163}{Transfer learning for
  low-resource neural machine translation}.
\newblock In {\em Proceedings of the 2016 Conference on Empirical Methods in
  Natural Language Processing\/}. Association for Computational Linguistics,
  Austin, Texas, pages 1568--1575.
\newblock
  \href{https://aclweb.org/anthology/D16-1163}{https://aclweb.org/anthology/D16-1163}.

\end{thebibliography}
\bibliographystyle{acl_natbib}

\end{document}